\documentclass{article}
\usepackage{amssymb}
\usepackage{amsmath}
\usepackage{graphicx}
\usepackage{hyperref}
\usepackage{wrapfig}
\usepackage{booktabs}
\usepackage{soul}
\usepackage{color}
\usepackage{subcaption}

\usepackage[preprint]{corl_2025} 

\title{Demystifying Diffusion Policies: Action Memorization and Simple Lookup Table Alternatives}

\author{
  Chengyang He\thanks{Equal contribution}\\
  National University of Singapore\\
  Singapore, Republic of Singapore \\
  \texttt{hecy@stanford.edu} \\
  \And
  Xu Liu\footnotemark[1]\\
  Stanford University \\
  California, United States \\
  \texttt{liuxujsw@stanford.edu} \\
  \And
  Gadiel Sznaier Camps \\
  Stanford University \\
  California, United States \\
  \texttt{gsznaier@stanford.edu} \\
  \And
  Guillaume Sartoretti \\
  National University of Singapore \\
  Singapore, Republic of Singapore \\
  \texttt{guillaume.sartoretti@nus.edu.sg} \\
  \And
  Mac Schwager \\
  Stanford University \\
  California, United States \\
  \texttt{schwager@stanford.edu} \\
}

\begin{document}
\maketitle


\begin{abstract}
    Diffusion policies have demonstrated remarkable dexterity and robustness in intricate, high-dimensional robot manipulation tasks, while training from a small number of demonstrations. However, the reason for this performance remains a mystery. In this paper, we offer a surprising hypothesis: diffusion policies essentially memorize an action lookup table---\emph{and this is beneficial}.  We posit that, at runtime, diffusion policies find the closest training image to the test image in a latent space, and recall the associated training action sequence, offering reactivity without the need for action generalization. This is effective in the sparse data regime, where there is not enough data density for the model to learn action generalization.  We support this claim with systematic empirical evidence.  Even when conditioned on wildly out of distribution (OOD) images of cats and dogs, the Diffusion Policy still outputs an action sequence from the training data. With this insight, we propose a simple policy, the Action Lookup Table (ALT), as a lightweight alternative to the Diffusion Policy. Our ALT policy uses a contrastive image encoder as a hash function to index the closest corresponding training action sequence, explicitly performing the computation that the Diffusion Policy implicitly learns.  We show empirically that for relatively small datasets, ALT matches the performance of a diffusion model, while requiring only 0.0034 of the inference time and 0.0085 of the memory footprint, allowing for much faster closed-loop inference with resource constrained robots.  We also train our ALT policy to give an explicit OOD flag when the distance between the runtime image is too far in the latent space from the training images, giving a simple but effective runtime monitor.
    More information can be found at: \url{https://stanfordmsl.github.io/alt/}.
\end{abstract}

\keywords{Robot Manipulation, Diffusion Policy, Action Memorization} 


\section{Introduction}
\label{sec:intro}

Imitation learning for robot manipulation requires training a policy to map from image inputs to action sequence outputs given a relatively small number of demonstrations.  Recently, the Diffusion Policy~\cite{chi2023diffusion} has emerged as a powerful and novel approach to this problem by modeling the robot’s visuomotor policy as a denoising diffusion probabilistic model \cite{ho2020denoising}.  Diffusion models are generative models that train a filter to iteratively remove noise from noise-corrupted training data.  At inference time a random output is sampled, and progressively denoised (conditioned on the input) to produce an inference.  The diffusion model was originally introduced for image generation and remains the dominant architecture in that domain~\cite{saharia2022photorealistic,stable_diffusion_wiki}. 
The primary advantage of the Diffusion Policy for robot maipulation lies in its ability to model multi-modal action distributions, scale to high-dimensional outputs, and produce long-horizon action sequences. 
Indeed, recent studies have shown that diffusion policies outperform many existing methods on challenging manipulation benchmarks~\cite{ze20243d,wang2024equivariant}.

The performance of the Diffusion Policy is unquestionable, however the explanation for this performance remains elusive.  In particular, typical diffusion policies are trained on 50--200 task demonstrations (a small amount of data), while maintaining the same number of parameters (typically over 100 million) as image generation models trained on billions of images \cite{saharia2022photorealistic,stable_diffusion_wiki}.  Furthermore, the common practice is to train Diffusion Policies until the training loss is low, but the test loss is high---the classic signal for over-fitting in machine learning.  In typical machine learning, overfitting is associated with poor test-time performance and poor generalization.  Yet it is observed that this overfitting is actually necessary for strong test time performance of the Diffusion Policy. The question arises: 

\textit{Why do diffusion policies trained to overfit small data sets appear to give strong test-time performance in robot manipulation?}  

In this paper we show that, indeed, diffusion policies severely overfit the training data, such that they memorize the training action sequences.  At inference, diffusion policies recall the training action sequences with nearly no generalization. They essentially perform a lookup table that maps runtime images to training action sequences. Combined with online closed-loop execution with runtime images, this action memorization appears to be a winning recipe for strong manipulation policies obtained from small amounts of demonstration data. The stochastic multi-modality of the policy results from slight randomness in which training action sequence is indexed, not from generalization over action sequences. However, one key drawback of the Diffusion Policy is slow inference time, which leads to slow robot execution punctuated by pauses as the model recomputes the next inference at the end of each action sequence. This motivates our second research question: 

\textit{Can the same action memorization behavior be accomplished with a simpler, faster model architecture to yield faster runtime performance?}

\begin{wrapfigure}{r}{0.5\textwidth}
  \vspace{-0.3cm}
  \centering
  \includegraphics[width=0.5\textwidth]{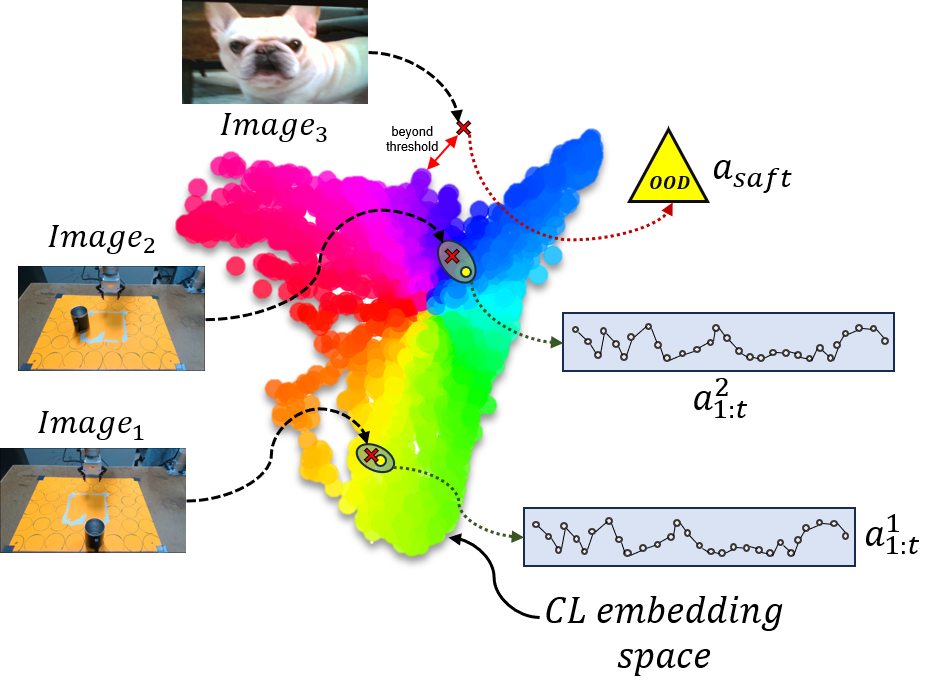}
  \vspace{-0.7cm}
  \caption{Resulting latent space of our constrastive learning (CL) based ALT on our training data, illustrating the distribution of training examples and example in- (1,2) and out-of-distribution (3) test images.}
  \label{fig:alt}
  \vspace{-10pt}
\end{wrapfigure}

To answer this question, we propose a simple lookup table policy with a trained image encoder to map from images to actions, which we call the Action Lookup Table (ALT) policy, illustrated in Fig.~\ref{fig:alt}.  The ALT policy performs similarly to the Diffusion Policy, while being 300 times faster at inference, and requiring less than 1/100th the memory footprint.

Concretely, we design a lightweight image-joint pose encoder that maps each provided observation into a low-dimensional feature representation (LFR), which we then use to create a memory bank of LFRs from the training expert demonstrations. 
During inference, we then encode new observations, find the nearest code from the training set in the latent space, and output the memorized training action sequence associated with that code. 
The enccoder is trained using a contrastive learning objective~\cite{chen2020simple}, encouraging positive sample pairs to be embedded closer together while pushing negative pairs apart in the latent space. ALT avoids the costly iterative denoising steps, enabling faster (single forward-pass) inference.

Our \textbf{contributions} are as follows: (1) We hypothesize the Diffusion Policy implicitly memorizes an action lookup table. We provide conceptual intuition for this hypothesis and support it through extensive empirical validation. (2) We present a new ALT policy~(as shown in Fig.~\ref{fig:alt}), a contrastive learning-based alternative to diffusion policies, explicitly implementing an action sequence lookup table by encoding image-joint pose inputs into a latent space to index training action sequences.  By setting a threshold distance in the latent space, we obtain a simple OOD detector to flag potential policy failures at runtime.


\section{Prior Work}
\label{sec:priorwork}
Diffusion models, trained by gradually adding Gaussian noise to data during training \cite{ho2020denoising, ramesh2021zero, blattmann2023stable}, were originally developed for high-dimensional data generation tasks such as image, video, or audio synthesis~\cite{nichol2021improved,rombach2022high}. These models can produce seemingly novel high-quality images and videos in a variety of different styles through simple text \cite{ramesh2021zero,ruiz2023dreambooth, balaji2022ediff,saharia2022photorealistic} and image \cite{saharia2022palette, tumanyan2023plug, ceylan2023pix2video} prompt conditioning. In order to capture the complex multimodal distributions inherent in visual and auditory data, these models are often large, containing from hundreds of millions to billions of parameters~\cite{saharia2022photorealistic,stable_diffusion_wiki}, and are trained over large datasets with hundreds of millions to billions of examples \cite{rombach2022high}. 

Leveraging the strong performance of diffusion models, the Diffusion Policy~\cite{chi2023diffusion} achieves state-of-the-art performance in visuomotor control for single skill imitation learning. Trained with a limited number of expert demonstrations, the model learns to predict a sequence of robot actions \cite{chi2023diffusion, ren2024diffusion, lu2024manicm, lee2024diff} conditioned on a given observation. This observation can be images \cite{chi2023diffusion}, point clouds \cite{ze20243d}, semantic labels \cite{wang2024gendp, li2024language} or potential fields \cite{mizuta2024cobl}. Due to its apparent robustness to perturbations, diffusion policies have been deployed for a wide range of robotics tasks, including manipulation \cite{black2023zero, kim2022diffusionclip, chi2023diffusion}, multi-skill learning \cite{chen2023playfusion, xu2023xskill}, and motion planning \cite{shaoul2024multi, serifi2024robot, sridhar2024nomad}. Diffusion models have also been used in robotics for data augmentation \cite{kapelyukh2023dall, chen2023diffusion} to aid in the training of other models.

The phenomenon of memorization in diffusion models has been well-studied in image generation, but not in robotics, to our knowledge. \cite{gu2023memorization} observed that smaller datasets are prone to cause memorization, especially when conditioned with uninformative labels, while \cite{somepalli2023diffusion} discovered that reconstructive memorization occurs even for models trained on enormous datasets, with as much as 2\% of the generated images being duplicates of the training data. Similarly, \cite{carlini2023extracting} demonstrated a way to extract known training examples from state-of-the-art models, such as DALL-E 2 \cite{ramesh2022hierarchical}. Meanwhile, \cite{gu2023memorization} notes that the traditional denoising score matching objective used during training has a closed-form optimal solution that can only replicate training images. In contrast, \cite{jain2024classifier} posits that the denoising process causes diffusion models to learn an attraction basin for each training sample, thereby guiding prompt-conditioned generated images towards memorized data. \cite{wen2024detecting} corroborates this by noting that diffusion models tend to converge to a known training sample regardless of initialization, suggesting memorization of both the prompt and the denoising trajectory. Similarly, both \cite{somepalli2023understanding} and \cite{chen2024extracting} note that, although less prevalent than conditioned models, memorization still occurs in unconditioned models and \cite{hintersdorf2024finding} finds that memorized data are often associated with corresponding individual neurons.

Modifying the loss function \cite{chen2024towards}, gradients \cite{chen2024towards}, conditioning approach \cite{jain2024classifier}, or keyword prompts \cite{wen2024detecting, somepalli2023understanding, ren2024unveiling} during training and inference are all typical methods for reducing model memorization. However, we propose that while memorization is undesirable for image generation due to privacy and copyright concerns, it is actually beneficial for robotics. When the diffusion model is used in domains with rich input space (e.g., images) but limited output space (e.g., robot actions), the gap between model capacity and output dimensionality, combined with the use of imitation learning (which inherently lacks task-level supervision), makes overfitting via memorization a plausible explanation for its strong performance in in-distribution settings.


\section{Diffusion Policy Analysis}
\label{sec:dp_analysis}

\subsection{Preliminaries}

The output, $\mathbf{x}^0$, of a diffusion model, $\varepsilon_\theta$, is obtained by iteratively removing noise (i.e. denoising) from a starting value, $\mathbf{x}^k$, sampled from a Normal Distribution, $\mathcal{N}(0,\sigma^2I)$. The denoising process evolves according to 
\begin{equation} \label{eq:sld}
    \begin{aligned}
        \mathbf{x}^{k-1} = \alpha(\mathbf{x}^k-\gamma\varepsilon_\theta(\mathbf{x}^k,k) + \mathcal{N}(0,\sigma^2I)
    \end{aligned}
\end{equation}
to remove the noise in $k$ steps based on a predetermined noise schedule that specifies the values of $\alpha$, $\sigma$, and $\gamma$ at each iteration. This procedure can be thought of as a single stochastic gradient descent step $x' = x-\gamma\nabla E(x)$, where the model $\varepsilon_\theta$ is used to predict the gradient field $\nabla E(x)$. A more detailed explanation of the denoising process can be found in \cite{chi2023diffusion} and \cite{ho2020denoising}.

\subsection{Diffusion Model Generalization Regimes}
\label{sec:dma}

\begin{figure}[t]
\centering
\begin{minipage}{0.24\textwidth}
    \centering
    \includegraphics[width=\linewidth, trim={120 120 120 130}, clip]{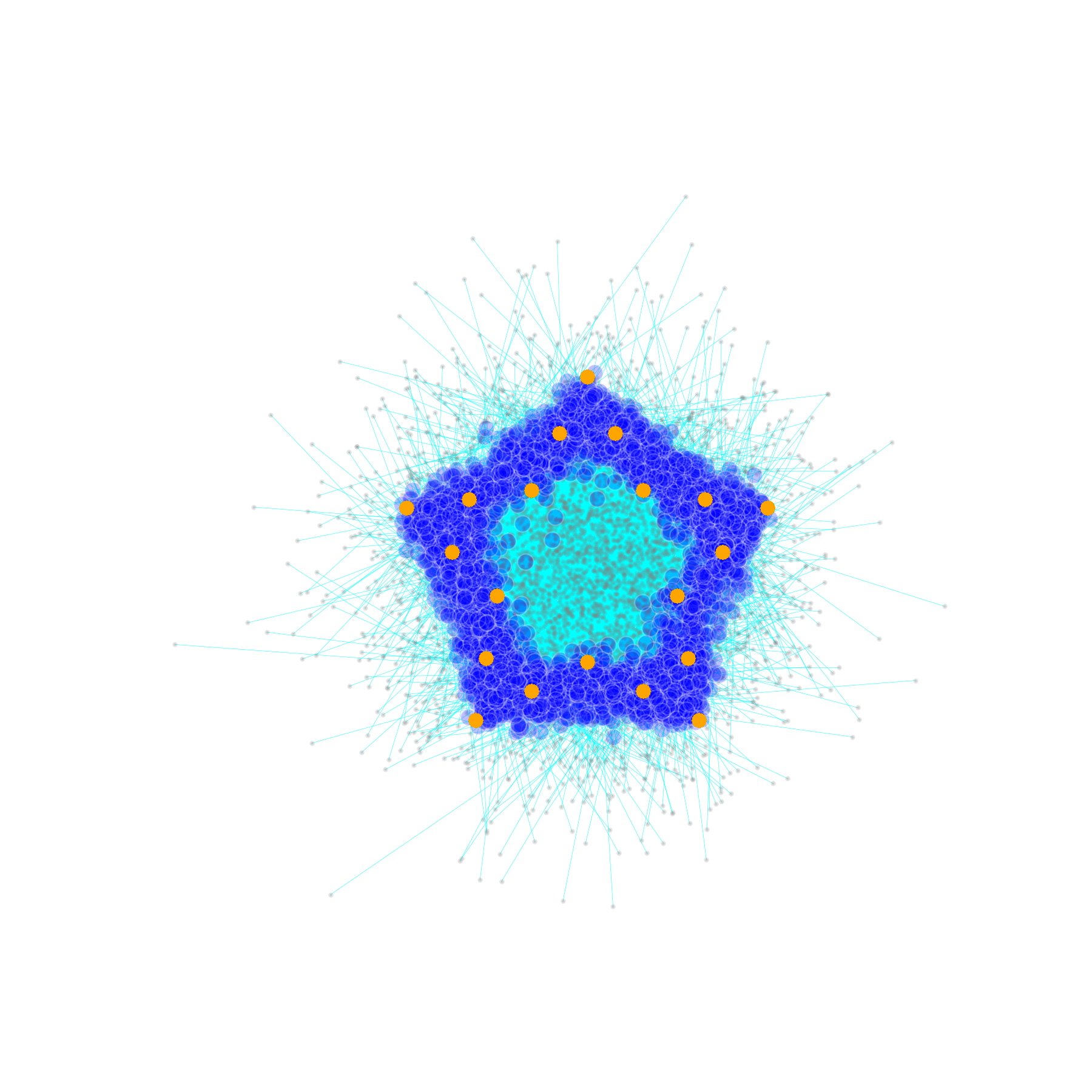}
    \captionsetup{justification=centering}
    \subcaption{low-capacity model, small data (SD)}
\end{minipage}
\hfill
\begin{minipage}{0.24\textwidth}
    \centering
    \includegraphics[width=\linewidth, trim={120 120 120 130}, clip]{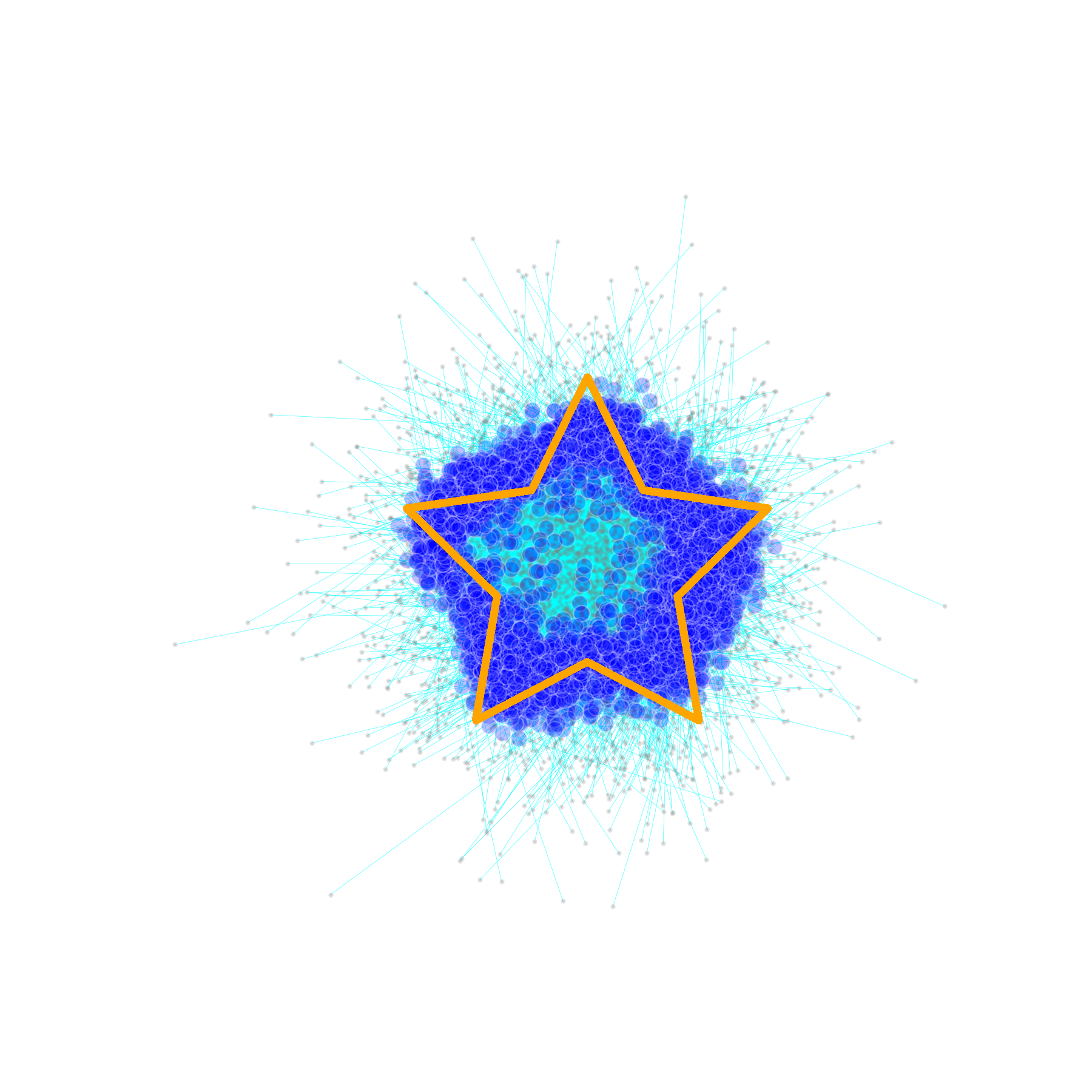}
    \captionsetup{justification=centering}
    \subcaption{low-capacity model, large data (LD)}
\end{minipage}
\hfill
\begin{minipage}{0.24\textwidth}
    \centering
    \includegraphics[width=\linewidth, trim={120 120 120 130}, clip]{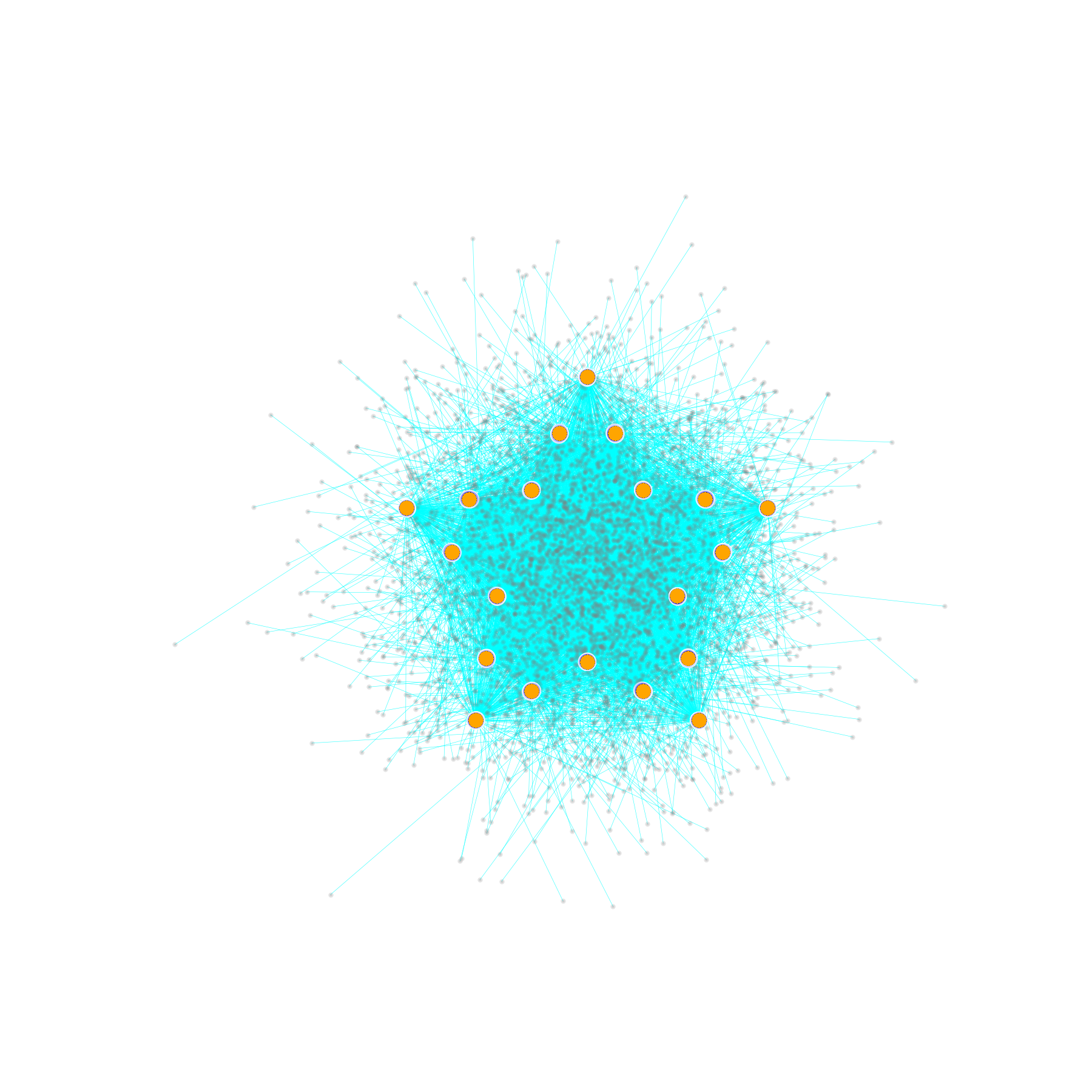}
    \captionsetup{justification=centering}
    \subcaption{high-capacity model, SD (e.g. Diffusion Policy)}
\end{minipage}
\hfill
\begin{minipage}{0.24\textwidth}
    \centering
    \includegraphics[width=\linewidth, trim={120 120 120 130}, clip]{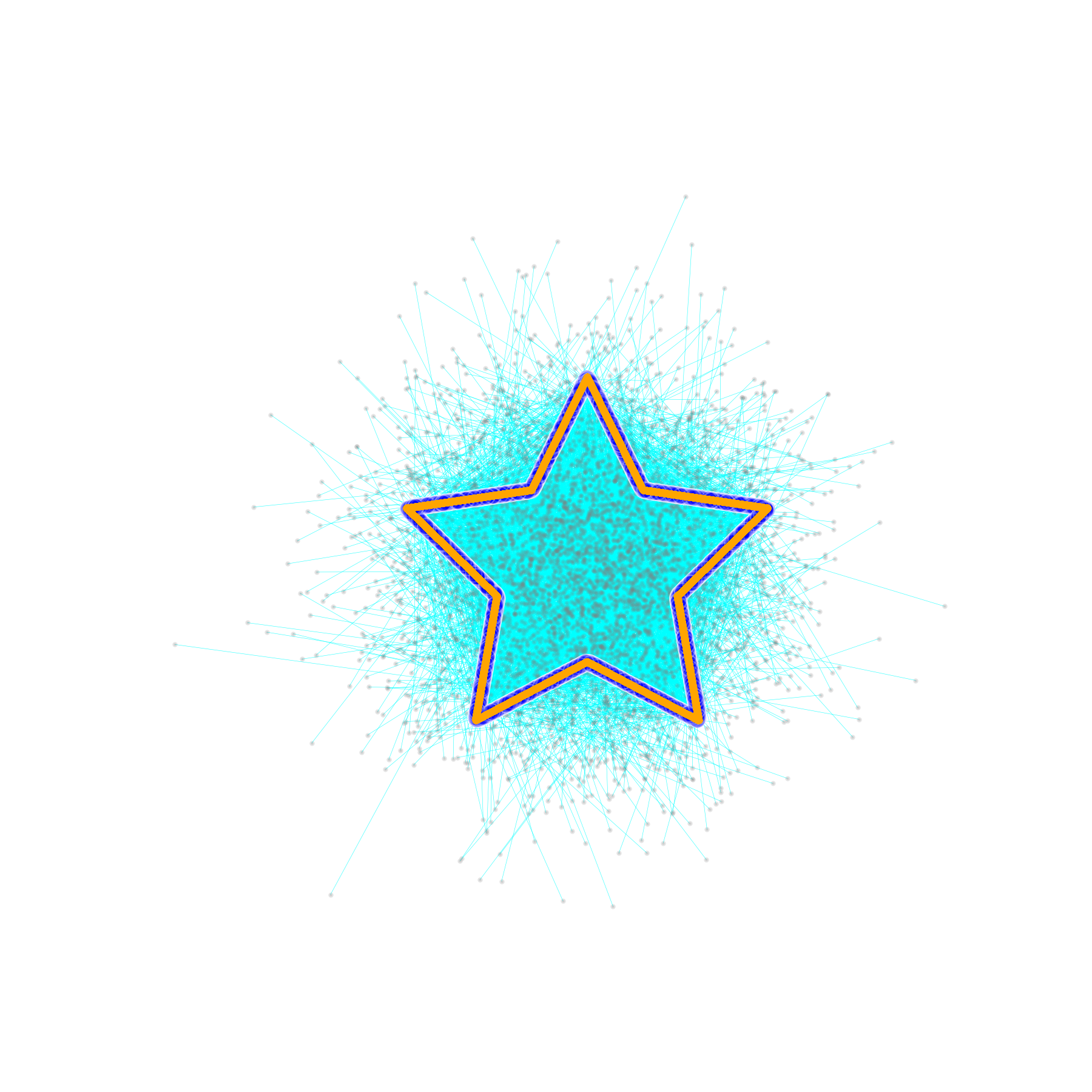}
    \captionsetup{justification=centering}
    \subcaption{high-capacity model, LD (e.g. Image Diffusion)}
\end{minipage}
\vspace{-0.15cm}
\caption{Training a generative model from 2D points uniformly distributed on a star-shaped 1D manifold. Orange indicates the training samples, black represents the random seeds used for diffusion, light cyan lines show the denoising flow direction, and blue marks the final inference results. Each subplot shows a different training regime: (a) A low-capacity model ($\sim$146 parameters) trained on a small dataset (20 samples) gives erratic inferences. (b) The low-capacity model trained on a large dataset (100k samples) generalizes to the wrong manifold. (c) A high-capacity model ($\sim$9.5 million parameters) trained on a small dataset (the Diffusion Policy regime) approximately memorizes the dataset, but does not generalize.  All the inference (blue) overlay the training data (orange) points, essentially implementing a lookup table. (d) A high-capacity model trained on a large dataset shows strong generalization to the correct data manifold (regime of large scale image diffusion models). Similar observations are demonstrated for other 2D manifolds in Appendix~\ref{appx:da}.}
\label{fig:dp_analysis}
\vspace{-0.4cm}
\end{figure} 

We illustrate four generalization regimes for a simple Multi-Layer Perceptron (MLP)-based diffusion denoising model trained to learn a ground-truth distribution consisting of 2D points uniformly sampled on a 1D manifold shaped as a star (Fig.~\ref{fig:dp_analysis}). 
We show qualitative model performance with a low-capacity vs high-capacity MPL, trained with small vs large data sets. As expected, when a low-capacity model is trained on a small dataset (Fig.~\ref{fig:dp_analysis}a), it fails to fit the data adequately. Similarly, due to its limited capacity, when such a model is given sufficient data (Fig.~\ref{fig:dp_analysis}b), it is only able to learn an approximation that oversimplifies the data manifold (here, approximating a star shape as a hexagon). In comparison, when a high-capacity model is trained on a small dataset (Fig.~\ref{fig:dp_analysis}c), the diffusion model tends to memorize the individual training samples rather than generalizing or interpolating between them. 
This memorization allows accurate fitting of the limited training points (good for robot manipulation tasks), but results in the model failing to capture the broader underlying data manifold, a behavior that is consistent with our findings for diffusion policies for robot manipulation. This phenomenon is related to manifold overfitting~\cite{loaiza2022diagnosing}: when a powerful generative model is trained on data lying on a narrow sub-manifold, it might fit the data too closely while struggling outside that sub-manifold.  When the model is provided sufficient data (Fig.~\ref{fig:dp_analysis}d), it is now able to effectively fit both the data and the true underlying distribution, representing the regime common in large scale image generation models.\footnote{More results can be found in Appendix~\ref{appx:da}} 
However, acquiring large-scale expert demonstrations for robot manipulation that evenly and densely cover the action sequence space remains a significant practical challenge. As Diffusion Policies are trained on larger and larger datasets, they may move toward the large data regime (Fig.~\ref{fig:dp_analysis}d)  with true generalization on the action manifold, but this seems to be beyond the current state of the art.


\subsection{Hypothesis and Experiments}

\begin{wrapfigure}{r}{0.7\textwidth}
  \vspace{-10pt}
  \centering
  \includegraphics[width=0.7\textwidth]{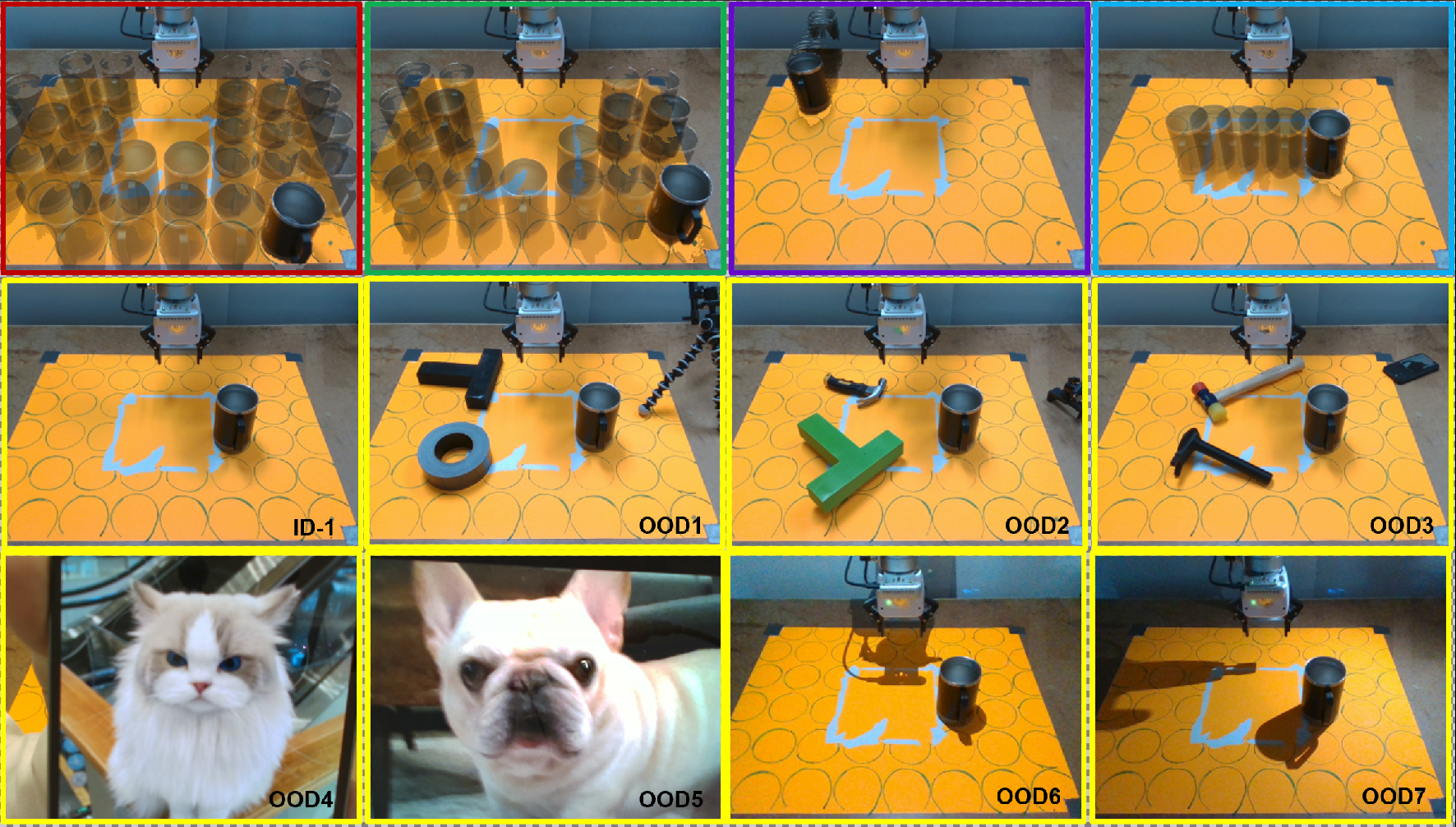}
  \caption{The first panel (red outline) shows in-distribution (InD) tests. The second panel (green outline) shows an InD interpolation test, with cup evenly placed between training positions. Panels with purple and blue outlines illustrate cases where the cup is gradually moved from an in-distribution location to new OOD position. The remaining panels (yellow outline) introduce OOD image distractors to assess the model’s robustness.}
  \label{fig:exp}
  \vspace{-10pt}
\end{wrapfigure}

To evaluate the hypothesis that the Diffusion Policy implicitly implements an action sequence lookup table, we designed a series of cup grasping experiments. We trained a diffusion policy for cup grasping using the standard codebase from \cite{chi2023diffusion}, trained with 30 demonstrations of cup locations evenly spaced throughout the workspace, with a held-out square in the middle, as indicated by the green circles and blue tape in Fig.~\ref{fig:exp}.  The robot has a third-person view fixed camera and a wrist mounted camera, both used to condition the policy.
For each position, we performed one demonstration (to remove the confounding effect of multi-modal action generation).
We then validated the learned policy on the 30 in-distribution cases, confirming its ability to reproduce the training demonstrations. 

To further investigate the action generalization behavior of the policy, we systematically introduced a variety of interpolation and extrapolation inputs, ranging from in-distribution (InD) to out-of-distribution (OOD) and analyzed the resulting behavior.
Specifically, we designed four scenarios: 
(1:InD-Interpolate) Placing the cup at evenly spaced test positions located between the original training positions (Fig.~\ref{fig:exp} green border); 
(2:OOD-Interpolate) Slowly moving the cup from one in-distribution position through an OOD region (blue tape square) to another in-distribution position (Fig.~\ref{fig:exp} blue border);
(3:OOD-Extrapolate) Gradually moving the cup from an in-distribution position to an OOD location outside the fixed camera's field of view (Fig.~\ref{fig:exp} purple border).
(4:OOD-Distractors) Introducing OOD visual distractors of varying difficulty levels (Fig.~\ref{fig:exp} yellow border), including wildly OOD images of a cat and a dog;
These settings allowed us to explore and analyze the generalization behavior and potential memory-driven characteristics of the Diffusion Policy. If the Diffusion Policy were performing action generalization, one would expect the following in each scenario: (1:InD-Interpolate) interpolation in the action space; (2:OOD-Interpolate) some action interpolation with degraded performance in the middle, where it is far from the training examples; (3:OOD-Extrapolate) progressively degraded action performance as the object moves farther from the training set; and (4:OOD-Distractors) degraded action performance as the number and severity of distractors grows, with dog and cat inducing erratic action sequences.  

\emph{In fact, all of these behavioral expectations are incorrect. In every case, the Diffusion Policy almost exactly reproduces one of the training action sequences} as explained below. This is consistent with our action lookup table hypothesis. 

\subsection{Results}

In this subsection, we introduce a custom metric, the similarity score, designed to quantify how closely an inference action sequence resembles sequences from the training set.
It is defined as: $\mathcal{S}=1-\frac{s(\tau^{(r)},\tau^{(1)})}{s(\tau^{(1)},\tau^{(2)})}$,
where $s(\tau^{(r)},\tau^{(1)})$ denotes the average Euclidean distance between the matched points on the current action sequence and its closest training sequence, and $s(\tau^{(1)},\tau^{(2)})$ denotes the distance between the second-closest and the closest training action sequence. 
We normalize the similarity score such that it is exactly $0.5$ if the inference trajectory interpolates half way between the two closest training trajectories, and is $1.0$ if it perfectly re-executes a training trajectory. 
If an action sequence closely follows a specific training sequence while maintaining a clear separation from other nearby sequences, this provides strong evidence of memory-based retrieval rather than action generalization. Note that this similarity metric does not measure action \emph{quality}, just action recall.  For example, the robot may take a highly ineffective action sequence, but if it closely matches one of the training sequences, the similarity score will be high.  We make no claims on the effectiveness of the Diffusion Policy actions.  Just that they are recalled from a memorization of the training actions.

\begin{figure*}[t]
  \centering
  \includegraphics[width=0.9\textwidth, trim={0 23 0 0}, clip]{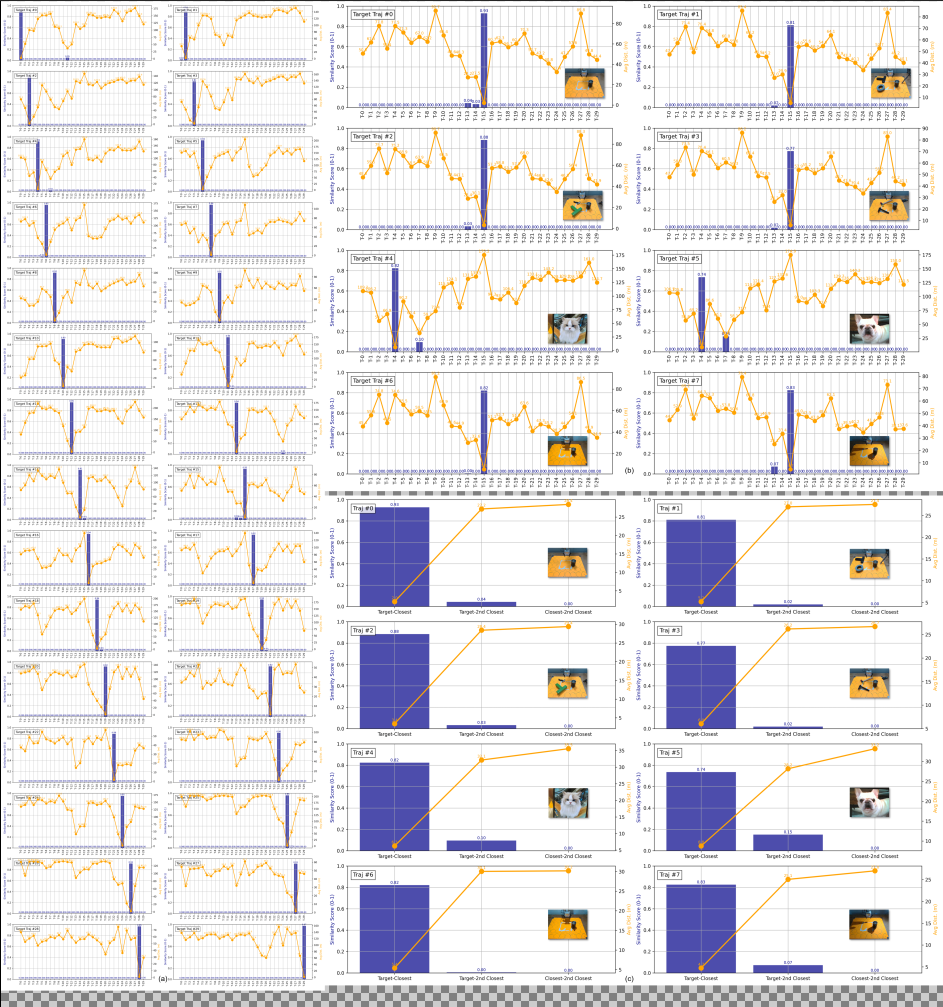}
  \caption{Similarity and distance statistics between inference and training trajectories. Each subplot shows the similarity scores (blue bars) and average distances (orange lines) between the Diffusion Policy inference and training trajectories. The large gap between the closest and second-closest neighbors indicates strong alignment with specific training examples.}
  \label{fig:analysis}
  \vspace{-10pt}
\end{figure*}

Fig.~\ref{fig:analysis} provides compelling evidence supporting the hypothesis that the Diffusion Policy exhibit memory-based action cloning behavior. 
We first validate this in the in-distribution setting, where cups are placed exactly as they were during training. 
As shown in Fig.~\ref{fig:analysis} (left), each action sequence almost perfectly overlaps with the corresponding training sequence. 
The distance to the nearest neighbor (yellow polyline) is near zero, while the distance to the second nearest trajectory is substantially larger, resulting in a similarity score close to $1$ (blue bar). 
This indicates that the model is essentially replaying an action sequence memorized during training when presented with familiar inputs.
This behavior persists even under OOD scenarios. 
In Fig.~\ref{fig:analysis} (top right) and \ref{fig:analysis} (bottom right), we introduce distractors to the environment (as shown in Fig.~\ref{fig:exp} yellow border). 
However, the model consistently follows the trajectory most similar to a training example.  
In Fig.~\ref{fig:analysis} (bottom right), regardless of whether distractors are present, the distance to the closest training trajectory remains very low, and the distance to the second closest remains high. 
As a result, the similarity score of the nearest neighbor remains close to $1.0$, while the second nearest similarity approaches $0.0$.  Again, we do not find that the action sequence is the ``right one'' in the face of distractors, but rather that the model chooses one of the training sequences to re-execute, even when the input image is OOD.

Fig.~\ref{fig:analysis} (top right) provides a global view of similarity scores across all training trajectories. 
In the presence of distractors, almost all high-similarity matches are sharply concentrated on a single training trajectory, indicating a surprising OOD default behavior.  The diffusion model seems to revert to one or two fallback action sequences when presented with OOD images. 
Even when the input is entirely unrelated to the task, for example, an image of a cat or a dog, the diffusion model still produces an action sequence that closely resembles one from the training set.
We believe those results show that the Diffusion Policy's decision-making is largely governed by memory retrieval, rather than by generalized reasoning over the action space.
In both clean and distractor-laden scenarios, the model demonstrates consistent action replay behavior, supporting our hypothesis that its decision-making is fundamentally memory-driven. 
Additional analyses and results also support this observation (for example, in the InD-Interpolate cases, the Diffusion Policy outputs a trajectory that closely matches one of the four corresponding nearest-neighbor training trajectories). 
See Appendix~\ref{appx:ar} for more OOD scenarios. 


\section{ALT: the Action Lookup Table Policy}
\label{sec:alternative}

Building upon these results, we design a lightweight alternative method while still achieving comparable functionality.
Our policy, functioning similarly to a hashing function, is based on contrastive learning and retrieves demonstration trajectories using an ALT mechanism (as shown in Fig.~\ref{fig:alt}) within the contrastive learned latent space.
If our hypothesis holds, this method should deliver performance on par with the Diffusion Policy, while also offering more predictable fallback behaviors in the presence of out-of-distribution (OOD) inputs, therefore improving safety and robustness.

\subsection{Training Phase}

\begin{figure*}[t]
\centering
\includegraphics[width=0.85\textwidth, trim={0 0 0 0}, clip]{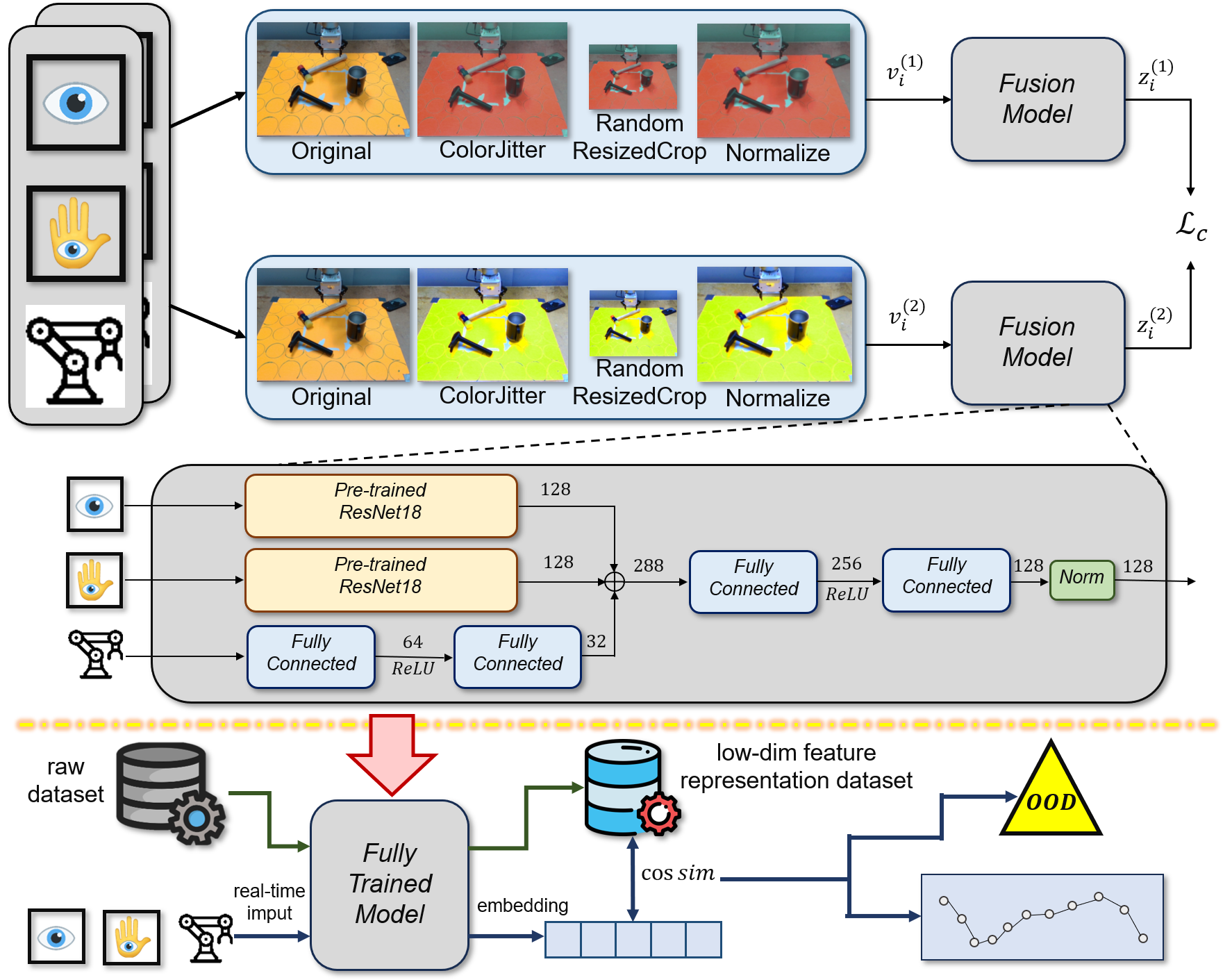}
\caption{Contrastive training (top, above the yellow dashed line) and inference (bottom) phases of our ALT policy. 
The inference process has two stages: the green arrows indicate building the ALT latent space with the trained model, while blue arrows represent real-time inference.
}
\label{fig:model}
\vspace{-0.4cm}
\end{figure*}

At each timestep, our data consists of three parts: a first-person view from the robot arm end-effector, a third-person view, and the end-effector pose denoted as $D=\{(I_i^h,I_i^t,p_i) \}^{N}_{i=1}$, where $I_i^h$ and $I_i^t$ are the hand camera image and third-view image respectively, and $p_i$ represents the position and orientation of the end-effector.
We designed a fusion encoder to integrate these inputs into a unified embedding for contrastive learning. 
The architecture of this model is illustrated in Fig.~\ref{fig:model}, and can be formulated as: 
\begin{equation}
    \begin{aligned}
        z_i = f_{fusion}(f_{img}(I_i^h),f_{img}(I_i^t),f_{pose}(p_i)).
    \end{aligned}
\end{equation}
For visual encoding $f_{img}$, we employ a pre-trained ResNet-18~\cite{he2016deep} as the image encoder backbone. 
Trained on large-scale datasets such as ImageNet, this pretrained network is capable of extracting general-purpose visual features. 
Leveraging such pre-trained features is beneficial in our setting, where the available dataset is relatively small.
To perform contrastive learning, we adopt the NT-Xent loss, which requires the fused embeddings to be L2-normalize $z_i\leftarrow \frac{z_i}{||z_i||_2}$. 
This ensures that the similarity computation in the embedding space is stable and consistent.

As shown in Fig.~\ref{fig:model}, we adopt a contrastive learning framework to extract robust and discriminative representations from each frame using alignment across multiple modalities.
We generate two different data augmentation views for each sample $d_i=(I_i^h,I_i^t,p_i)$. 
Specifically, we apply a composed image augmentation pipeline $\mathcal{A}_1$ and $\mathcal{A}_2$ that transforms each input $d_i$ into augmented version views $v_i^{(1)}$ and $v_i^{(2)}$. 
We feed $v_i^{(1)}$ and $v_i^{(2)}$ into the fusion encoder to obtain their embeddings $z_i^{(1)}$ and $z_i^{(2)}$. 
These two embeddings form a positive pair, and we train the network using the normalized temperature-scaled cross-entropy (NT-Xent) loss~\cite{chen2020simple} as our contrastive loss function:
\begin{equation}
    \begin{aligned}
        \mathcal{L}_{c} = -\frac{1}{2B}\sum_{i=1}^B[\log\frac{\exp{(\text{sim}(z_i^{(1)},z_i^{(2)})/\tau)}}{\sum_{k\neq i}^{2B}\exp{(\text{sim}(z_i^{(1)},z_k)/\tau)}} + (1\leftrightarrow 2)],
    \end{aligned}
\end{equation}
where $\text{sim}(\cdot,\cdot)$ denotes the cosine similarity, and both inputs are L2-normalized prior to computation.
The parameter $\tau$ is the temperature (set as 0.4 in practice), which controls the sharpness of the similarity distribution, effectively scaling the logits to adjust the contrastive loss sensitivity.

\subsection{Inference Phase}

After training the model, we build a low-dimensional feature representation latent space to support trajectory matching and prediction during real-time execution. 
An overview of the full process is shown in Fig.~\ref{fig:model}.
Specifically, we need to encode each frame in the raw trajectory dataset using our fusion encoder, and the resulting embedding tensors are stored as entries in the database.
Another important component of the low-dimensional feature representation database is the trajectory ID and local frame index corresponding to each input.

During real-time inference, the incoming observation, consisting of the current third-person view, first-person view, and end-effector pose, is encoded into an embedding using the same fusion encoder. 
We then perform a search for best cosine similarity against the latent contrastive learning space to find the most similar stored embedding.
If the maximum similarity falls below a predefined threshold $\gamma$, the input is considered OOD, and the robot executes a safe fallback behavior when the input deviates significantly from the training distribution.
Otherwise, the system retrieves the matched trajectory ID and frame index, enabling real-time trajectory prediction and policy execution based on the stored demonstrations.

\subsection{Results}
In this section, we conduct two experiments to demonstrate the effectiveness of our proposed ALT policy and its action memorization mechanism. 
First, we test the model to assess its ability to enable successful task execution on a real robot under in-distribution conditions. Next, we assess the model’s performance under OOD conditions by introducing various distractors into the environment. 
As shown in Fig.~\ref{fig:exp}, we create OOD scenarios by placing additional unseen objects (e.g., tape, hammer), altering lighting conditions to produce varying shadows, or replacing the third-person viewpoint with entirely task-irrelevant images (such as pictures of a cat and a dog), while maintaining the cup in its original training position.
We compared our method with both a KD-Tree-based nearest neighbor retrieval and the Diffusion Policy to validate the feasibility of our explicit action memorization mechanism and demonstrate the advantages of our ALT policy. 
The results are summarized in Table~\ref{tab:result}, where a checkmark (\checkmark) indicates a successful match with the correct training action sequence, while a cross ($\times$) indicates an incorrect match. 
Because our method explicitly detects OOD cases, a green OOD denotes cases where the input is correctly identified as OOD yet the trajectory match remains accurate, demonstrating robustness. 
The red OOD indicates that an OOD input has been detected, and using such an input would result in an incorrect trajectory output.

\begin{table}[!t]
  \centering
  \caption{Experimental Results. 
  ID and OOD refer to in-distribution and out-of-distribution cases, respectively. 
  The first column reports the success rate of the policy in retrieving the correct training trajectory given the observations. 
  The second column shows the success rate of the policy in completing the task during a real-robot rollout. 
  The remaining column headers refer to scenarios shown in Fig.~\ref{fig:exp}. The MIT column denotes the \textbf{M}odel \textbf{I}nference \textbf{T}ime required for each method in seconds.}
  \scalebox{0.65}{
    \begin{tabular}{c|cccccccccc|c}
    \toprule              
    \textbf{Methods} & \textbf{Recall} &  \textbf{IDs} & \textbf{ID-1}   & \textbf{OOD1} & \textbf{OOD2}  & \textbf{OOD3}  & \textbf{OOD4}  & \textbf{OOD5} & \textbf{OOD6}  & \textbf{OOD7}  & \textbf{MIT}   \\
    \midrule\midrule                                 
    K-D Tree & 100\% & 63.3\% & \checkmark   & \checkmark   & \checkmark  & \checkmark   & \checkmark & $\times$  & \checkmark     & \checkmark  & $\sim$0.09 \\
    \midrule                              
    Diffusion Policy & 100\% & 100\% & \checkmark   & \checkmark  &  \checkmark  &  \checkmark  &  $\times$  &  $\times$  &  \checkmark  &  \checkmark  & $\sim$2.65 \\
    \midrule                             
    Ours w/ p, $\gamma=0.9$ & 100\% & - & \checkmark   & \checkmark  &  \color{green}OOD  &  \color{green}OOD  &  \color{red}OOD  &  \color{red}OOD &  \color{green}OOD  &  \checkmark  & $\sim$0.009 \\
    \midrule 
    Ours w/o p, $\gamma=0.9$ & 100\%& - & \checkmark   & \color{green}OOD  &  \color{green}OOD  &  \color{green}OOD  &  \color{green}OOD  &  \color{green}OOD 
    &  \color{green}OOD  &  \color{red}OOD  & $\sim$0.009   \\
    \midrule 
    Ours w/o p, $\gamma=0.75$ & 100\% & 100\% & \checkmark   & \checkmark  &  \checkmark  &  \checkmark  &  \checkmark  &  \checkmark 
    &  \color{green}OOD  &  \color{red}OOD & $\sim$0.009   \\
    \midrule 
    \bottomrule
\end{tabular}
  }
\vspace{-0.6cm}
  \label{tab:result}
\end{table}

It can be observed that when the amount of training data is relatively limited, all methods achieve perfect trajectory matching performance (100\%) in all in-distribution cases, demonstrating the effectiveness of the underlying memorization mechanism. 
In OOD scenarios, the KD-Tree method is able to correctly retrieve trajectories in most cases due to its exhaustive nearest neighbor search. 
However, because it performs pixel-wise comparisons between the input and the raw dataset, its worst-case computational complexity scales as $\mathcal{O}(N\cdot d)$, where $N$ is the number of stored trajectories and $d$ is the data dimensionality. 
This leads to high computational cost and inference time.
In comparison, the Diffusion Policy performs well in most scenarios but fails under more severe distribution shifts, such as in OOD4 and OOD5. Furthermore, due to the architectural complexity of diffusion models, its model inference time (MIT) is relatively long.
In contrast, our proposed ALT policy achieves comparable performance with a significantly smaller model size (45.5 MB vs. 5.3 GB for the Diffusion Policy), and a much faster inference time. 
While our approach also involves a complexity of $\mathcal{O}(N\cdot d)$, the embedding dimensionality $d$ is much smaller, which leads to a drastic reduction in computation time.


\section{Conclusion}
\label{sec:conclusion}

In this paper, we propose and validate a counter-intuitive hypothesis explaining the impressive performance of diffusion policies. 
Rather than generalizing actions, diffusion policies essentially memorize training actions through severe overfitting, effectively acting as implicit action lookup tables.
Our systematic experiments reveal that, even when confronted with unseen InD or completely OOD inputs, diffusion policies reliably reproduce memorized action sequences rather than generalizing beyond their training demonstrations.
Guided by this insight, we introduce a lightweight alternative, the ALT policy, explicitly utilizing a contrastive image encoder as a hash function to retrieve stored trajectories and action indices from memory.
When using a relatively small dataset, our ALT policy achieves performance comparable to the the Diffusion Policy while requiring only a fraction (0.34\%) of its inference time and substantially reduced memory usage (0.85\%), making it particularly suitable for deployment on resource-constrained robots.



\section{Limitations}
\label{appx:limits}

We acknowledge several limitations in our hypothesis and method. 
All current experiments and validations have been conducted on small-scale dataset, and it remains uncertain whether the performance can be maintained at the same level when applied to a larger dataset. 
For example, while KD-Tree exhibits strong matching performance in small datasets, its computational efficiency may degrade significantly as the number of entries increases. 
Similarly, although our ALT method shows clear advantages in terms of computational complexity, its scalability may also be dependent on the scale of training data, unlike the Diffusion Policy (whose model size is fixed).
Such behaviors still need to be fully explored in our future works. 
Lastly, another limitation of our approach is its sensitivity to hyperparameters, such as the number of training epochs, the temperature parameter in the contrastive loss, and the OOD detection threshold. 
We observed that careful tuning was required to achieve optimal performance.

\bibliography{reference}  

\newpage
\appendix

\section{Data Collection Pipeline}
\label{appx:pipeline}

\subsection{Robot Arm Data Collection}
\label{subsec:priordata}

Choosing an efficient, cheap, and safe method for data collection, crucial for robot imitation learning, remains an open problem. One of the most common solutions is to use remote controllers, such as VR, 3D space mouse, or smartphones.
However, due to high latency and indirect operation, the data collected in this way is often messy and low-quality, making it difficult to accurately capture human skills.
Fully synchronized systems with human operators, such as ALOHA \cite{zhao2023learning} and GELLO \cite{wu2024gello}, can solve this problem by allowing humans to teleoperate the robot in a more intuitive way while tracking the actions of this system in real time.
But, these methods require an additional specialized puppeting system, which incurs an additional cost. 
In comparison, UMI-gripper \cite{chi2024universal} is a cheap, intuitive, and robot-agnostic solution for data collection.
Yet, it cannot be used in our work as it is incompatible with situations where third-person perspectives are necessary, and limits the robot to a single manipulator that requires an expensive hardware interface. Thus, to collect the necessary data, we utilized a motion capture data collection method, MoDA, to capture high-quality action sequences with low latency.

\subsection{Motion-captured Demonstration for Arms}
\begin{figure}[t]
\centering
\includegraphics[width=5.4in]{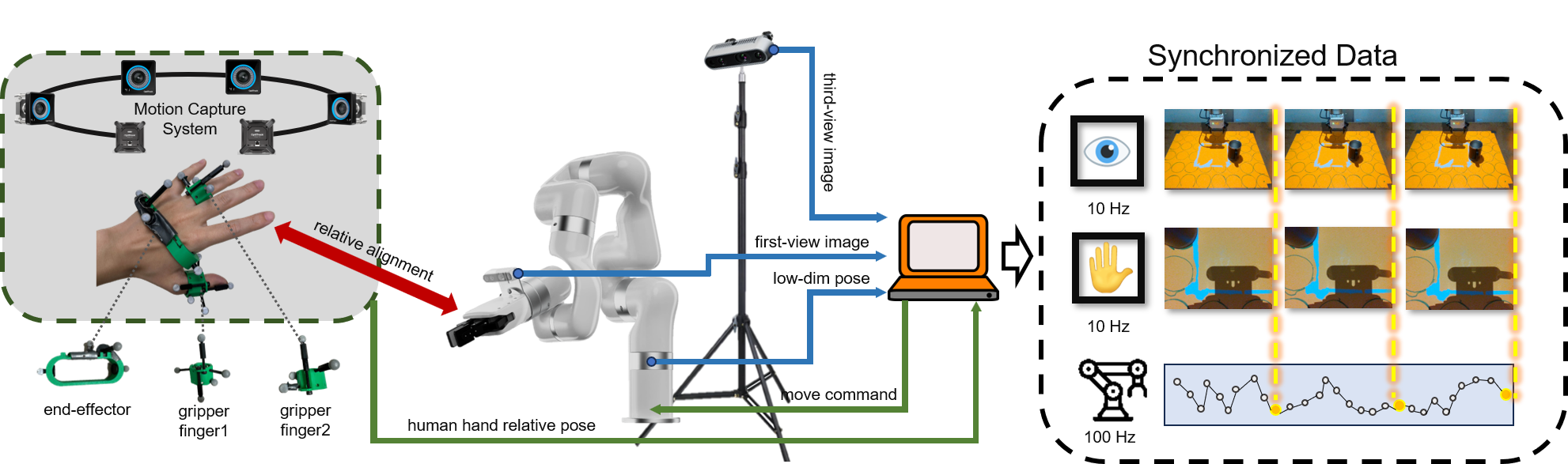}
\vspace{-0.3cm}
\caption{The data collection pipeline of MoDA (Motion-captured Demonstration for Arms). The green arrows indicate the process of aligning the relative positions of human's hand and the robotic arm, and the blue arrows indicate the data of collecting the robotic arm.}
\label{fig:pipeline}
\vspace{-0.4cm}
\end{figure}

Data collection plays a critical role in imitation learning, as the quality and generalizability of the learned policy depends heavily on the fidelity of the demonstrations.
In this work, we introduce MoDA (Motion-captured Demonstration for Arms), a streamlined and cost-effective data collection pipeline built upon motion capture (MoCap) systems that are commonly available in robotics laboratories (see Fig.~\ref{fig:pipeline}).
This pipeline provides high-fidelity human demonstrations for the robot, where a human demonstrator performs the cup grasping motion while wearing specialized trackers, and the system translates these motions into corresponding joint targets for a 6-DoF robot arm. 
MoDA can be extended to any other robot arm system with almost negligible cost, because our data collection pipeline is both task-agnostic and robot-agnostic.
To collect the necessary expert training demonstration data, we use an OptiTrack system to track the 6-DoF pose of the human palm in real time and map it directly to the end-effector of a robotic arm.
Simultaneously, we estimate the inter-finger distance to control the opening and closing of the gripper, thereby allowing us to signal when to grasp the cup. We then synchronize these actions with the corresponding in-hand and 3rd person camera views.
Compared to systems such as ALOHA, which rely on specialized and expensive teleoperation interfaces, our method does not need any active electronics or specialized wearables. Instead, the setup requires only a few 3D-printed brackets to attach passive IR reflective markers to the palm and fingers, making it an extremely low-cost and accessible solution when a MoCap system is already available in the lab. Furthermore, unlike UMI Gripper, which requires direct human interaction during data collection, our setup allows human demonstrators to operate out of frame, thereby ensuring clean third-person video demonstrations.
Compared to systems such as ALOHA, which rely on external equipment like teleoperation interfaces or instrumented gloves, our approach avoids the need for expensive or specialized hardware. 
In contrast to the UMI Gripper generated data, which often involves complex scenes with human demonstrators visibly present in the frame, our setup enables the collection of clean third-person video demonstrations where human demonstrators are minimally visible. 
This is particularly beneficial for training diffusion policies, as it minimizes noise and ambiguity in both the action and visual observation spaces, reducing the risk of learning failures due to poor-quality data.
In summary, unlike alternative setups that rely on specialized grippers, force sensors, or teleoperation rigs, our system can be assembled in-house with minimal resources and negligible additional expense.
Moreover, MoDA is not only task-agnostic, but also robot-agnostic, it does not rely on any specific type or model of robotic arm, making it highly adaptable across different hardware platforms and manipulation scenarios. 
This flexibility enables seamless integration into a wide range of experimental setups with minimal modification.

\section{Early Stopping Experiment}

\begin{figure}[htb]
\centering
\includegraphics[width=5.4in]{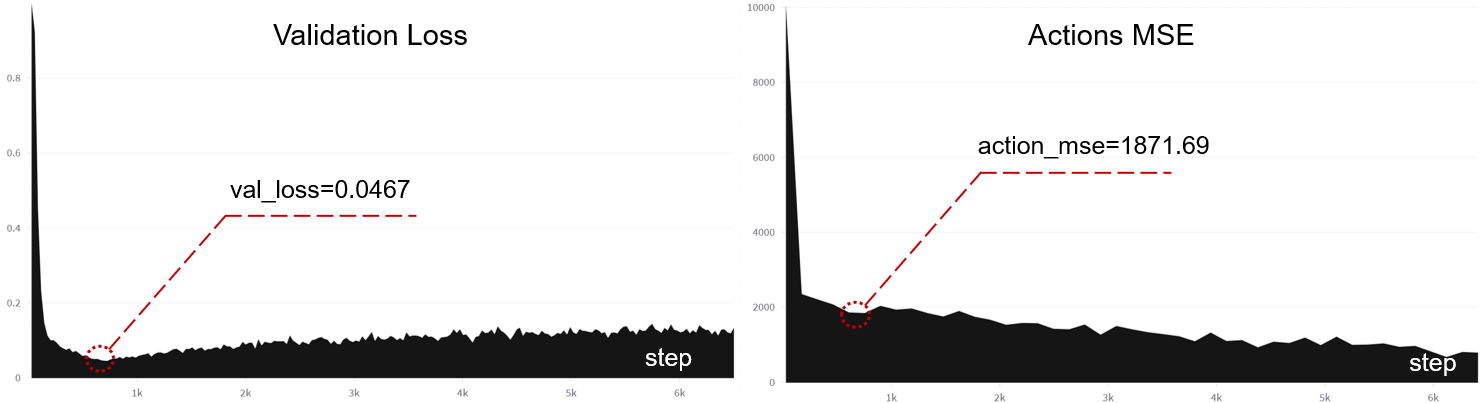}
\vspace{-0.3cm}
\caption{Early Stopping Experiment. The validation loss (left) reaches its minimum around step 650 before beginning to rise, indicating the onset of overfitting. 
However, at this point, the training action MSE (right) has not yet converged and remains as high as 1800. This suggests that more extensive training is necessary for the Diffusion Policy to output effective actions, even the input is in-distribution.}
\label{fig:early_stp}
\vspace{-0.4cm}
\end{figure}

To further support our hypothesis, we conducted an early stopping experiment. 
Early stopping is a common technique used to prevent potential overfitting during training, with the goal of improving a model’s generalization ability. 
In this experiment, we reserved one-third of the dataset as a validation set and used the remaining two-thirds for training. 
During training, we recorded both the validation loss and the mean squared error (MSE) between the predicted actions and ground-truth actions on the training set.
The first metric, validation loss, is used to determine when to stop training, thus preserving the model version with the best generalization. 
The second metric, actions MSE on training set, is used to monitor the model’s performance on the training set. 
As shown in Fig.~\ref{fig:early_stp}, although the validation loss reaches its minimum at a certain point, the corresponding action MSE remains high, around 1800. 
This result indicates that overfitting a diffusion policy model to the training data is a necessary requirement for producing accurate trajectories, as choosing the best model (chosen based on the validation loss) results in a policy that cannot reproduce the correct in-distribution trajectories.

\section{Diffusion Mechanism Analysis}
\label{appx:da}

\begin{figure}[htp]
\centering
\includegraphics[width=5.4in]{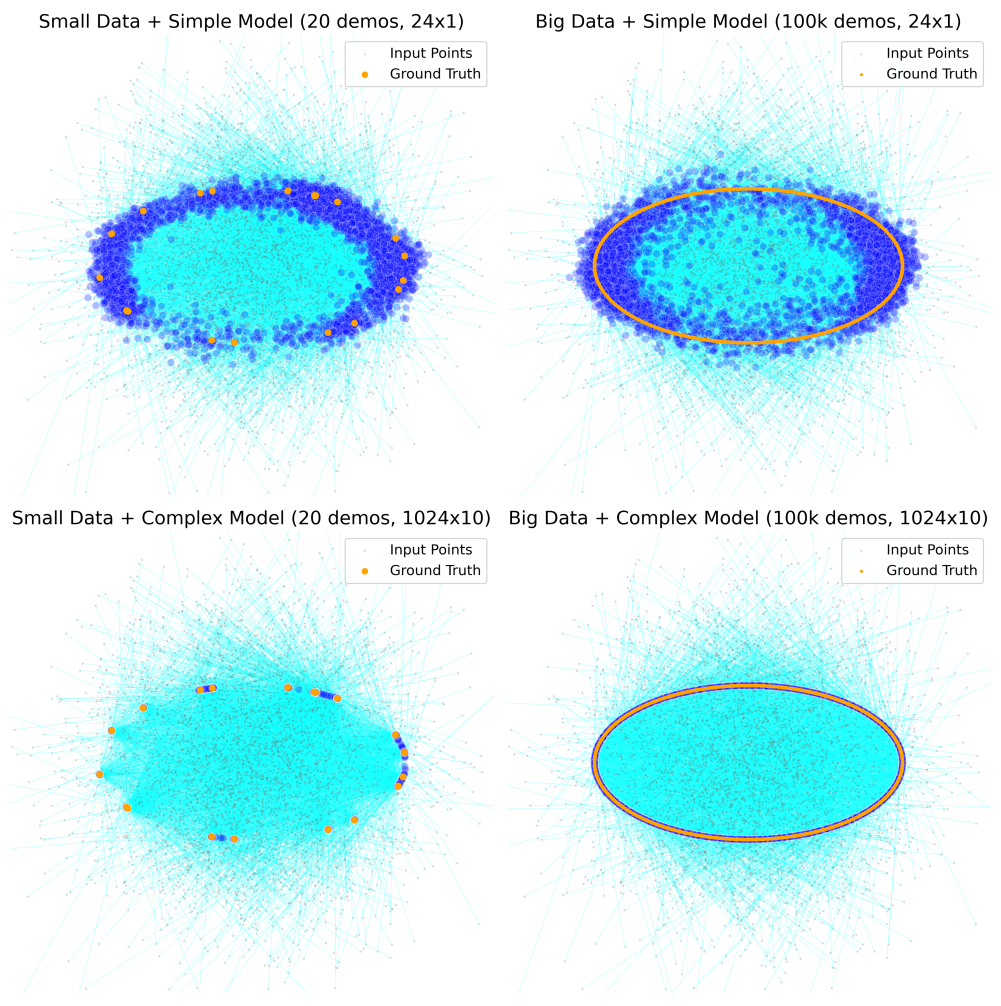}
\vspace{-0.3cm}
\caption{Training a generative model from 2D points on a ellipse-shaped 1D manifold. Orange points indicate training samples, gray points are noisy inputs, blue points are denoised outputs, and cyan lines shows the denoising directions.} 
\label{fig:appx_ellopse}
\vspace{-0.4cm}
\end{figure}

\begin{figure}[htp]
\centering
\includegraphics[width=5.4in]{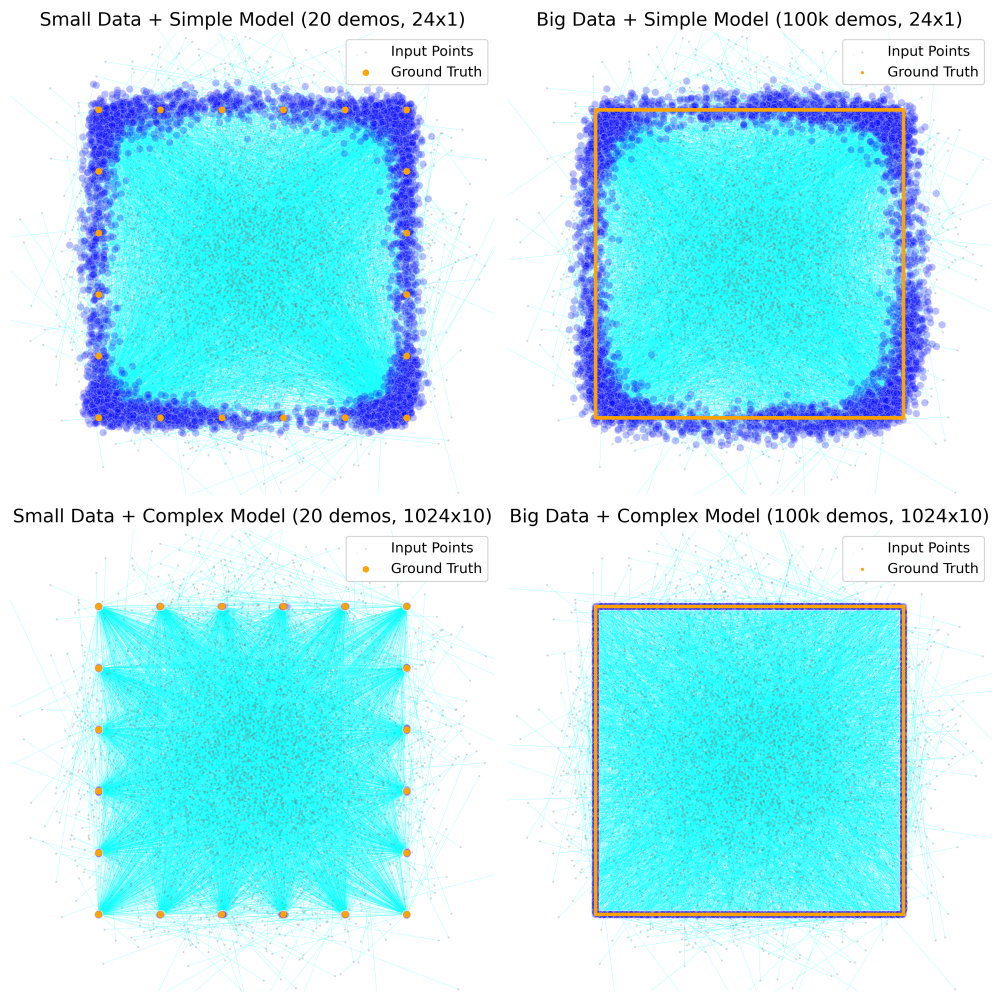}
\vspace{-0.3cm}
\caption{Training a generative model from 2D points on a rectangle-shaped 1D manifold.}
\label{fig:appx_rect}
\vspace{-0.4cm}
\end{figure}

\begin{figure}[htp]
\centering
\includegraphics[width=5.4in]{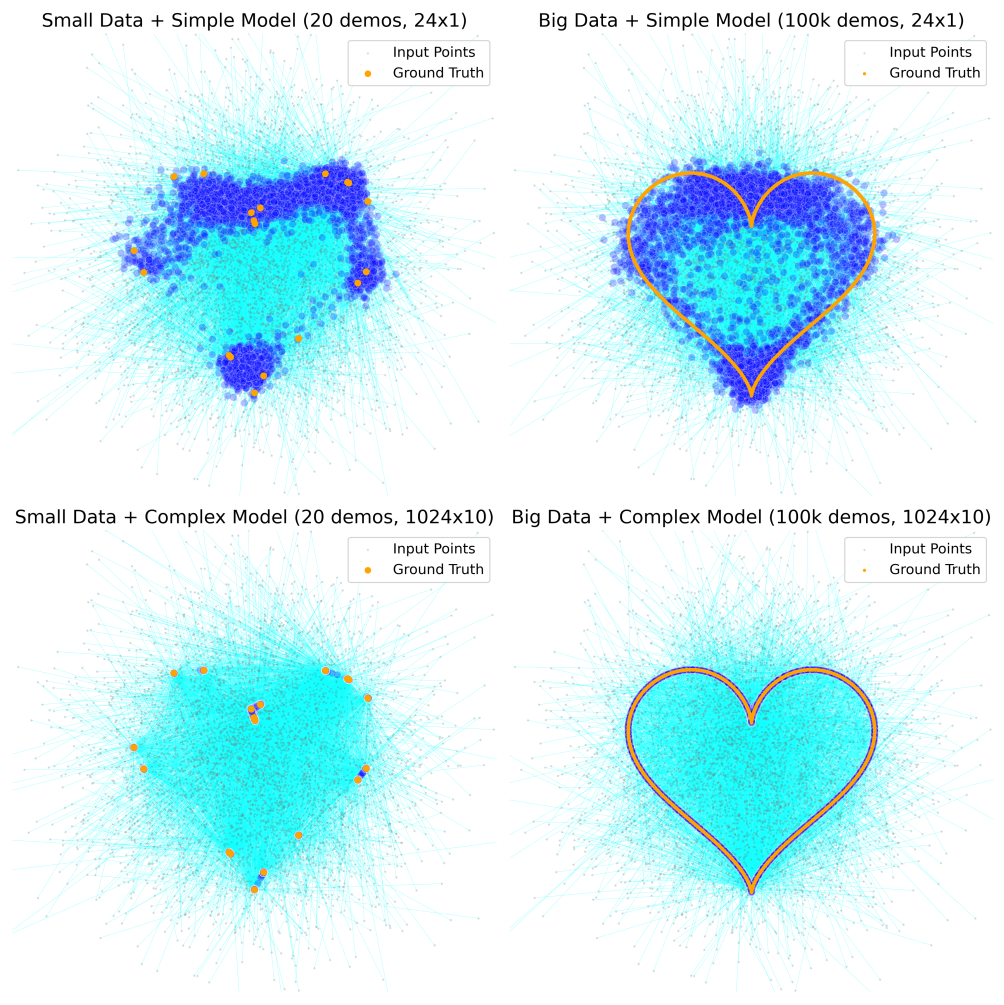}
\vspace{-0.3cm}
\caption{Training a generative model from 2D points on a heart-shaped 1D manifold.}
\label{fig:appx_heart}
\vspace{-0.4cm}
\end{figure}

In this section, we present additional examples to further illustrate the behavior of diffusion models under varying model capacities and dataset scales, as discussed in Section~\ref{sec:dma}. Specifically, we examine three additional 2D manifolds:  an ellipse~\ref{fig:appx_ellopse}, a rectangle~\ref{fig:appx_rect} and a heart shape~\ref{fig:appx_heart}. 

Consistent with the observations from Section~\ref{sec:dma}, low-capacity models trained on small datasets fail to accurately reconstruct manifolds, often producing noisy or collapsed outputs. 
Even when trained on large datasets, these models are limited by their representational capacity: simple shapes like ellipses can be approximated reasonably well (albeit still worse than with high-capacity models), more complex structures suffer significant distortion. For example, due to limited model expressiveness, low-capacity models approximate the heart shape as a crude triangle and smooth out the sharp corners of the rectangle.

Memorization behaviors are obvious when high-capacity models are trained on small datasets. 
Interestingly, although global structure reconstruction fails, local smoothness can still emerge. 
For instance, while the overall manifold may not be recovered, segments of an ellipse can still be accurately captured. 
This suggests that memorization in diffusion models is not absolute: when the training data is locally dense, models may still interpolate between nearby points, preserving some local structure. 
However, when larger gaps exist between segments, interpolation fails, and the model instead memorizes discrete samples without capturing the broader underlying manifold. 
When high-capacity models are instead trained on large datasets, all manifolds are accurately reconstructed, almost perfectly matching the true geometry. 
In practice, though, acquiring such large-scale expert demonstrations that sufficiently cover the state space remains a significant challenge in robotic manipulation tasks.

\section{Additional Results and Analysis}
\label{appx:ar}

\begin{figure}[t]
\centering
\includegraphics[width=5.4in]{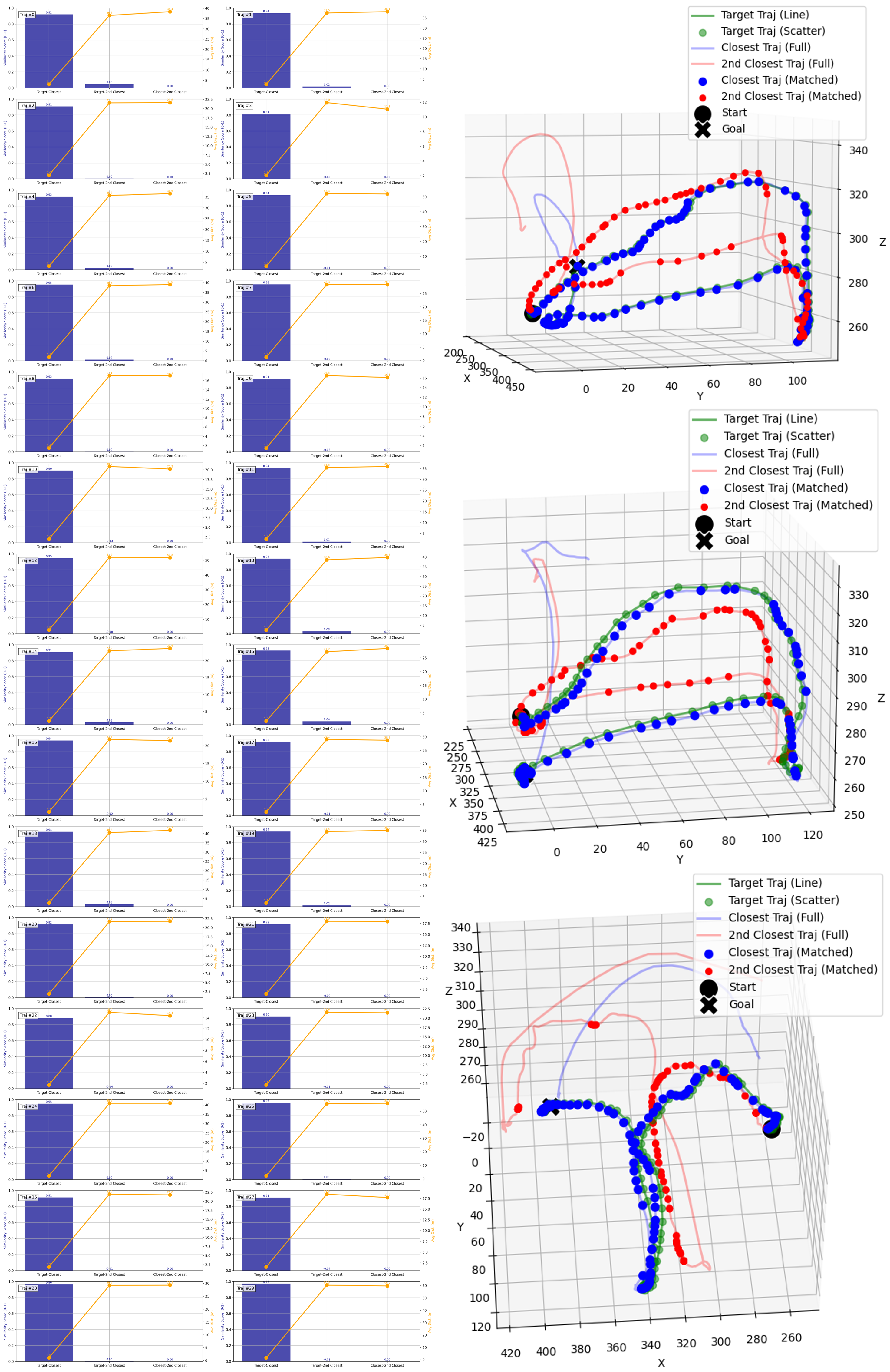}
\vspace{-0.3cm}
\caption{The Diffusion Policy inference result analysis under in-distribution conditions. On the left, each subplot shows the similarity distribution between the query inference trajectory and all stored trajectories in the database. On the right, the three figures provide 3D visualizations of representative matching cases. The green line represents the inference trajectory, while the blue and red dots show the closest and second-closest trajectories retrieved from the training set, respectively. }
\label{fig:appx_idx}
\vspace{-0.4cm}
\end{figure}

Fig.~\ref{fig:appx_idx} presents a supplementary visualization of the trajectory matching results under in-distribution conditions. 
In the left panels, for each case, the first blue bar represents the similarity score of the most closely matched training trajectory, while the second blue bar the similarity score of second closest trajectory. The consistently high top-1 similarity scores, combined with significant gaps to the second closest match, indicate clear and confident retrieval from the training data.

As shown in the right panels, the closest trajectory (blue) in the training dataset almost perfectly overlaps with the inference trajectory in all examples, showing that the Diffusion Policy can accurately retrieve the correct demonstration when the input remains within the training distribution.
These results strongly support our hypothesis that the Diffusion Policy depends on a memory-based retrieval mechanism to achieve its compelling results. 
The sharp similarity peaks and trajectory overlaps provide strong evidence that the model is not merely approximating the behavior, but is explicitly recalling memorized training trajectories under in-distribution conditions.

\begin{figure}[htb]
\centering
\includegraphics[width=5.4in]{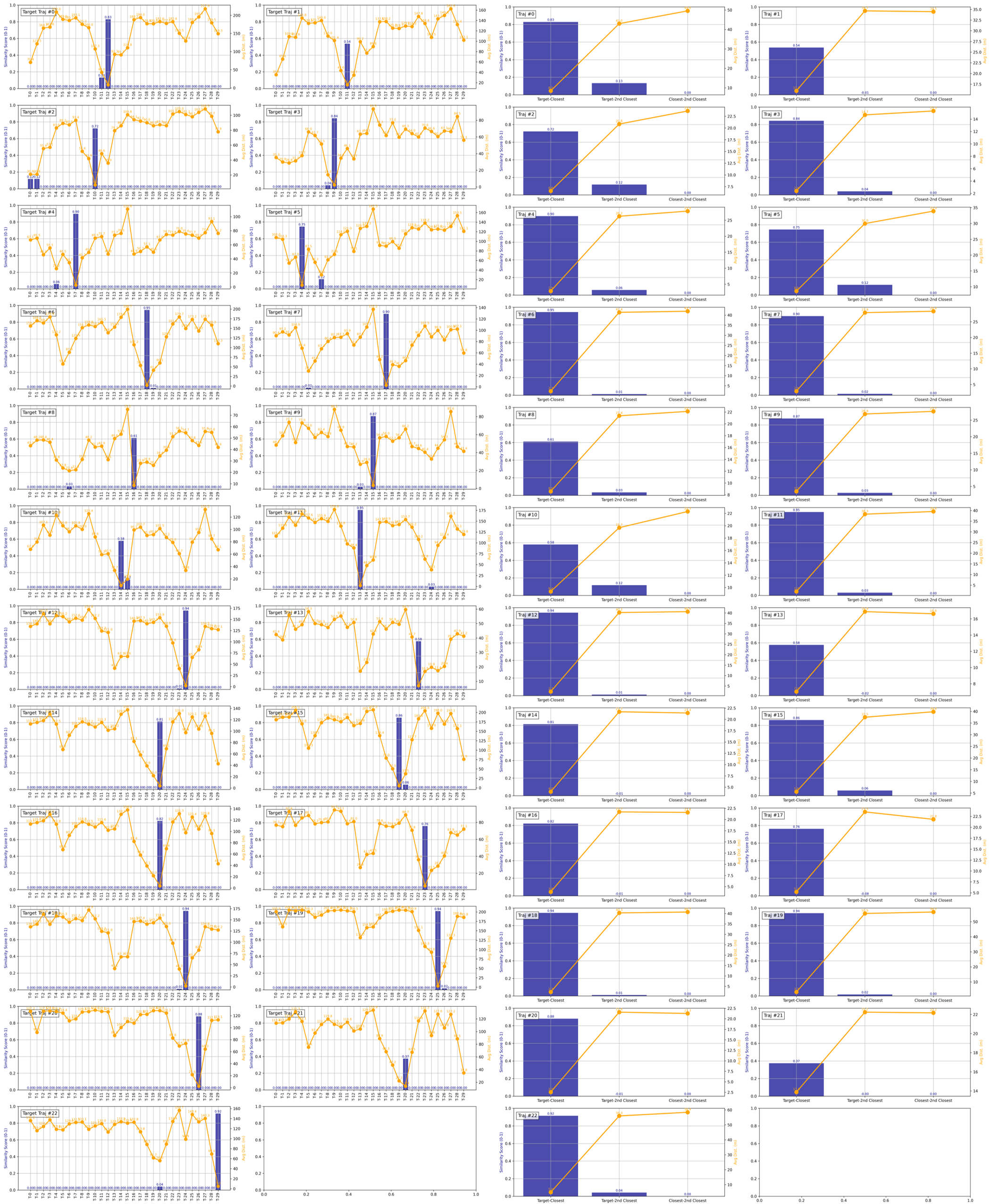}
\vspace{-0.3cm}
\caption{Inference trajectory analysis when the cup is placed between multiple In-distribution positions.}
\label{fig:appx_ood_evenly}
\vspace{-0.4cm}
\end{figure}

\begin{figure}[htb]
\centering
\includegraphics[width=5.4in]{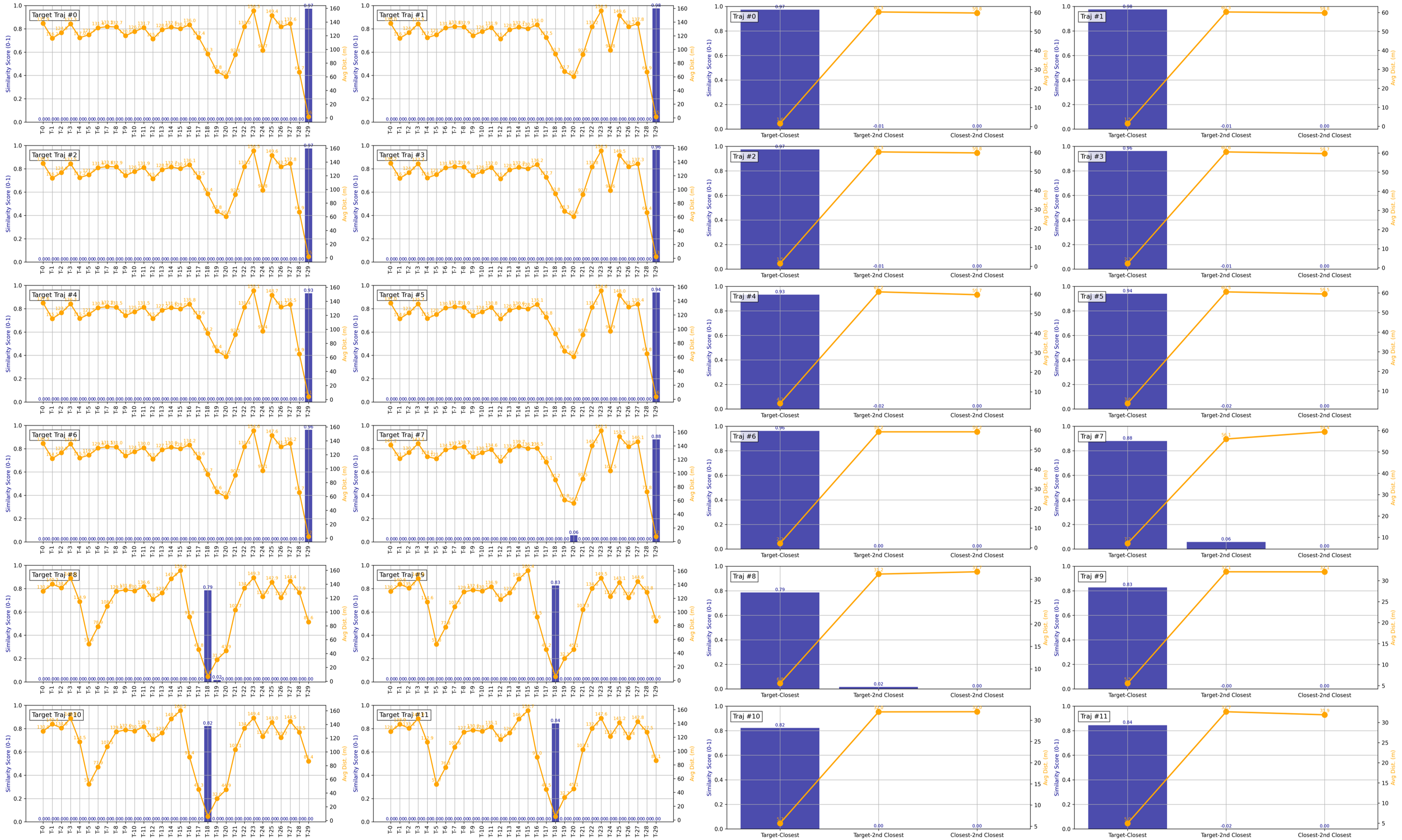}
\vspace{-0.3cm}
\caption{Inference trajectory analysis when the cup is gradually moved out of view from an in-distribution boundary.}
\label{fig:appx_ood_grad}
\vspace{-0.4cm}
\end{figure}

\begin{figure}[htb]
\centering
\includegraphics[width=5.4in]{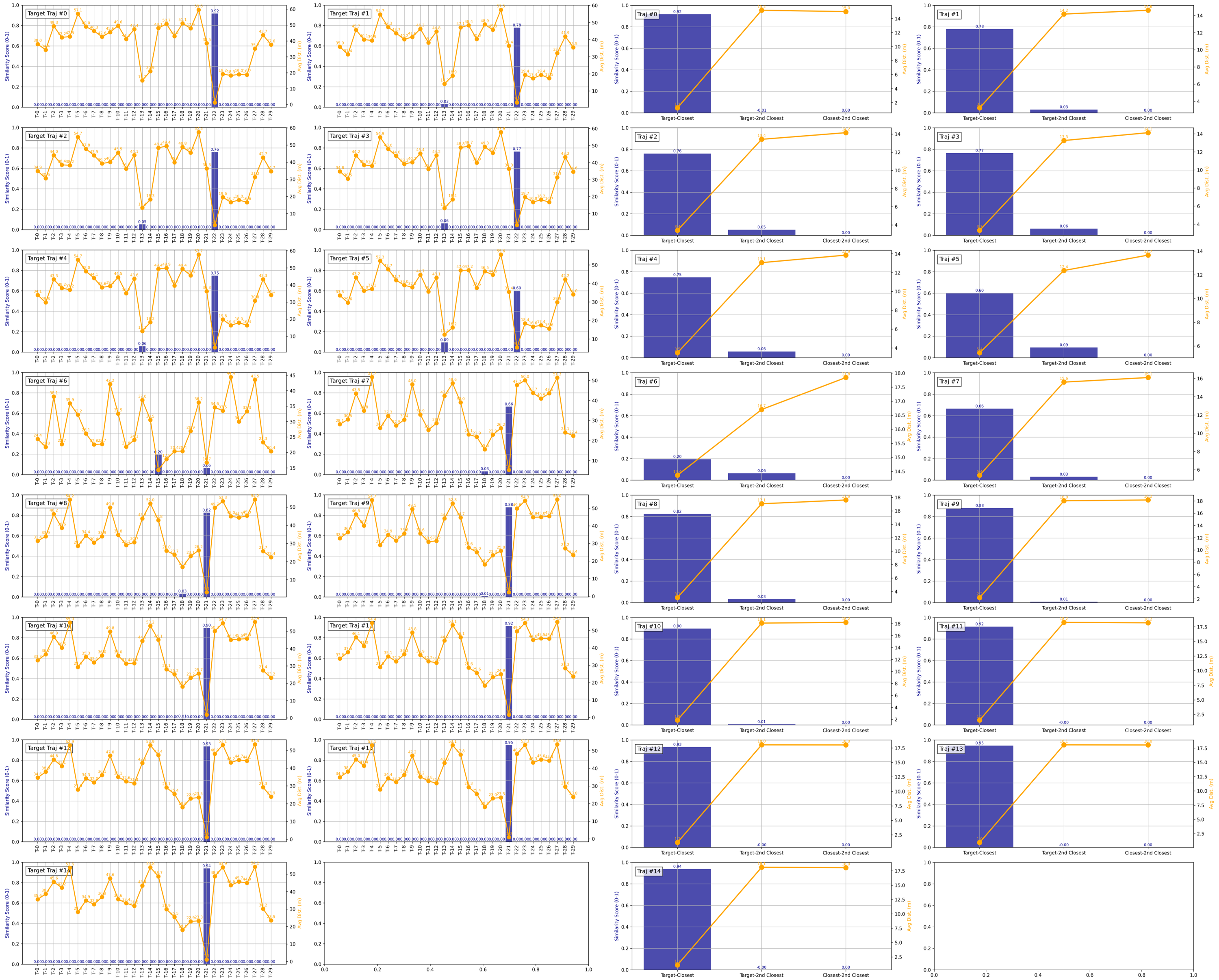}
\vspace{-0.3cm}
\caption{Inference trajectory analysis when the cup is gradually moved between two distant in-distribution positions.}
\label{fig:appx_ood_mid}
\vspace{-0.4cm}
\end{figure}

We further analyzed several additional out-of-distribution (OOD) scenarios, including: placing the cup evenly between three or four in-distribution positions (as shown in Fig.~\ref{fig:appx_ood_evenly}), gradually moving the cup out of the field of view from the edge of an in-distribution position (Fig.~\ref{fig:appx_ood_grad}), and slowly transitioning the cup between two distant in-distribution positions (Fig.~\ref{fig:appx_ood_mid}). 
It is evident that under these OOD conditions, the Diffusion Policy continues to produce trajectories that closely resemble those seen during training. 
These experimental results provide further support for our core hypothesis regarding the memory-driven behavior of diffusion policies.

\end{document}